\documentclass[3p,number]{elsarticle}
\usepackage{graphicx}
\usepackage{amsmath,amssymb} 
\usepackage{color}
\usepackage{amsmath,mathtools}
\usepackage[table]{xcolor}
\usepackage{multirow}
\usepackage{soul}
\usepackage{subcaption}
\usepackage[numbers]{natbib} 
\usepackage{epstopdf}
\usepackage{lineno}
\usepackage[hidelinks]{hyperref}
\usepackage{arydshln} 

\newcommand{\christy}[1]{\textcolor{black}{#1}}
\newcommand{\js}[1]{\textcolor{black}{#1}}

\biboptions{sort&compress}

\begin{document}

\begin{frontmatter}
\title{Spontaneous Subtle \christy{Expression} Detection and Recognition based on\\ Facial Strain
}
  \author[add1,add2]{Sze-Teng Liong\corref{cor1}}
  \ead{szeteng@siswa.um.edu.my}
  \author[add3]{John~See}
  \ead{johnsee@mmu.edu.my}
  \author[add2]{Raphael C.-W. Phan}
  \ead{raphael@mmu.edu.my}
  \author[add2]{Yee-Hui Oh}
  \ead{yeehui716@gmail.com}
  \author[add2]{Anh Cat Le Ngo}
  \ead{lengoanhcat@gmail.com}
  \author[add1]{KokSheik Wong}
  \ead{koksheik@um.edu.my}
  \author[add2]{Su-Wei Tan}
  \ead{swtan@mmu.edu.my}
  
  \cortext[cor1]{Corresponding author}
  \address[add1]{Faculty of Computer Science and Information Technology, University of Malaya, 50603 Kuala Lumpur, Malaysia}
  \address[add2]{Faculty of Engineering, Multimedia University, 63100 Cyberjaya, Malaysia}
  \address[add3]{Faculty of Computing and Informatics, Multimedia University, 63100 Cyberjaya, Malaysia}
  
\begin{abstract}
Optical strain is an extension of optical flow that is capable of quantifying subtle changes on faces and representing the minute facial motion intensities at the pixel level. This is computationally essential for the relatively new field of spontaneous micro-expression, where subtle \christy{expression}s can be technically challenging to pinpoint. In this paper, we present a novel method for detecting and recognizing micro-expressions by utilizing facial optical strain magnitudes to construct optical strain features and optical strain weighted features. The two sets of features are then concatenated to form the resultant feature histogram. Experiments were performed on the CASME II and SMIC databases. We demonstrate on both databases, the usefulness of optical strain information and more importantly, that our best approaches are able to outperform the original baseline results for both detection and recognition tasks. A comparison of the proposed method with other existing spatio-temporal feature extraction approaches is also presented.
\end{abstract}


\end{frontmatter}


\section{Introduction}
Micro-expression is one of the nonverbal communications that only occurs for a fraction of a second~\citep{nonverbal}. 
It is an uncontrollable expression that reveals the true emotional state of a person even when she is trying to conceal it. The appearance of a micro-expression is extremely rapid and brief, and it usually lasts for merely one twenty-fifth to one fifth of a second~\citep{duration}. This is the main reason why ordinary people always face difficulties in recognizing and understanding the genuine emotions of each other during real time conversations. There are six basic facial \christy{expressions}, notably happiness, surprise, anger, sad, fear and disgust, a categorization first proposed by \cite{six}.
Recognition of facial micro-expressions is valuable to various  applications in the field of medical diagnosis~\citep{feel}, national safety~\citep{security} and police interrogation~\citep{police}.The task of automatic recognition of spontaneous subtle \christy{expressions} is essentially of great interest to affective computing in this day and age. To date, many techniques and algorithms had been proposed and implemented for normal facial expression (or \emph{macro}-expression) detection and recognition \christy{\citep{ucar2016new,jia2015multi, zeng2006spon, ghi2016facial, wang2016facial}} but analysis on \emph{micro}-expression is still a relatively new research topic and very few works have been published \christy{\citep{liu2016main, wang2015micro, wang2014micro, wang2014face, huang2015facial}}.  

In one of our previous works~\citep{cv4ac} on subtle expression recognition, we proposed a technique that outperforms the baseline method of both CASME II~\citep{casme2} and SMIC databases~\citep{smic}, using optical strain magnitudes as weight matrices to improve the importance of the feature values extracted by Local Binary Patterns with Three Orthogonal Planes (LBP-TOP) in different block regions. In our second paper~\citep{liong2014optical}, recognition of micro-expressions was achieved using another technique. While only tested on the SMIC database, this method worked reasonably well by directly utilizing the optical strain features, following the temporal sum pooling and filtering processes. We further extend these two works, with substantial improvements and a more comprehensive evaluation on both detection and recognition tasks. 

In this paper, we introduce a novel method for automatic detection and recognition of spontaneous facial micro-expressions using optical strain information. \emph{Detection} refers to the presence of micro-expressions on the face without identification of its type, whereas \emph{recognition} goes a step further to distinguish the exact state or type of expression shown on the face. The proposed method mainly builds on the feature extraction process the optical flow method proposed by~\cite{robust}, which gives rise to the notion of optical strain. The feature histogram is constructed using optical strain information, following three main processes: (1) All the optical strain images in each video are temporally pooled, then the strain magnitudes of the pooled image are treated as features; (2) Optical strain magnitudes are pooled in both spatial and temporal directions to
form a 
weighting matrix. The respective weights of each video are then multiplied with features from the $XY$-plane extracted by LBP-TOP; (3) Lastly, the feature histograms from processes (1) and (2) are concatenated to form the final resultant feature histogram of the video sample.

The rest of the paper is organized as follows. Section~\ref{sec:related} briefly reviews the related work. Section~\ref{sec:fe} describes the spatio-temporal features used in the paper, followed by Section \ref{sec:proposed} that explains the proposed algorithm in detail. The description of the databases used are discussed in Section~\ref{sec:experiment}. The experiment results for detection and recognition of micro-expressions are summarized in Section~\ref{sec:results}. Finally, conclusion is drawn in Section~\ref{sec:conclusion}.

\section{Related Work}
\label{sec:related}
In the paper by \cite{towards}, optical strain pattern was used for spotting facial micro-expressions automatically. They achieved 100\% detection accuracy on the USF micro-expression database, \christy{with one false spot}. After computing the strain values for each pixel of the frames, a local threshold value was calculated by segmenting each frame into three pre-defined regions (i.e., forehead, cheek and mouth). The optical strain magnitudes that fell within certain threshold boundaries were considered a macro-expression. If the high strain values detected in less than two facial regions and only lasted for less than 20 frames (the frame rate was 30$fps$), it only considered to be a micro-expression. 
%
However, the dataset used possess a small sample size, and contains a total of only 7 micro-expressions. Besides, the micro-expressions detected were not spontaneous but rather posed ones, which are less natural and realistic.

The same authors~\citep{longvideo} later carried out an extensive test on two larger datasets 
containing a total of 124 micro-expressions. They also implemented an improved algorithm to spot the micro-expressions on these two datasets. Instead of partitioning the frame into three regions, they divided each frame into eight regions, namely: forehead, left and right of eye, left and right of cheek, left and right of mouth and chin. Some parts of the face were masked to overcome the existing noise problem in the captured image frames. A promising detection accuracy of 74\% was achieved. 


LBP-TOP \citep{dynamic} 
describes the space-time texture of a video volume, which encodes the local texture pattern by thresholding the center pixel against its neighbouring pixels. Block-based LBP-TOP partitions the three orthogonal image planes into $N\times N$ non-overlapping blocks, where the final histogram is a concatenation of histograms from each block volume. This final histogram represents the appearance, horizontal motion and vertical motion of a video. 
It has been a growing interest in micro-expression recognition recently, with a majority of works \citep{wang2014micro,davison2014micro} employing LBP-TOP and other variants \citep{wang2015efficient} as their choice of feature.

The large number of pixels in an image or video can be summarized into a more compact lower dimension representation. Generally, feature pooling in spatial domain is one commonly employed technique, which partitions the image into several regions, then summing up or averaging the pixel intensities of each region. Spatial pooling was employed together with state-of-the-art feature descriptors such as SIFT~\citep{sift} and Histograms of Oriented Gradients (HOG)~\citep{hog} to enhance their robustness against noise and clutter \citep{boureau}. Different variations of pooling functions (e.g., maximum, minimum, mean and variance) can also be applied to summarize features over time to improve the performance of an automatic annotation and ranking music radio system~\citep{music}.

There are many facial expression databases widely used in literature~\citep{survey}. However, spontaneous micro-expressions databases that are publicly available are somewhat limited, which is one of the most challenging obstacle towards the work of automatic recognition of micro-expressions. Naturally, this can be attributed to the difficulties faced in proper elicitation and manual labeling of micro-expression data. In addition, there remain several flaws with the existing micro-expression databases that hinders the progress of this research. For instance, USF-HD~\citep{usf} and Polikovsky's database~\citep{poli} 
%
recorded only 100 and 42 videos respectively, both containing posed micro-expressions instead of spontaneous (i.e., natural) ones. Since micro-expressions are typically involuntary and uncontrollable, theoretically they cannot be imitated or acted out~\citep{life}. Hence, the spontaneity of these micro-expressions is an essential characteristic for realistic and meaningful usages. Interestingly, the USF-HD also uses an unconventional criteria for determining micro-expressions (i.e., 2/3 second), which is longer than the most accepted durations. 

Another database, YorkDDT~\citep{york} is a spontaneous micro-expression database 
which collected for a deception detection test (DDT) as part of a psychological study. Although consists of spontaneously obtained samples, its inadequacy lies in its rather short samples, low frame rate (i.e., 25 \emph{fps}), while the original videos also had no expression labels.
Irrelevant head and face movements are obvious when the subjects are speaking, contributing to more hurdles in the aspect of face alignment and registration for a recognition task. Similarly, spontaneous micro-expressions were also used in the Canal-9 political debate corpus~\citep{longvideo}, although the reported detection performance was only as good as a chance accuracy (i.e., 50\%). This was attributed to the head movement and talking, while a much larger set of samples could also provide better generalization of patterns.

\section{Feature Extraction}
\label{sec:fe}
\subsection{Optical Flow}
Optical flow is a popular method of estimating the image motion between two successive frames and it is expressed in a two-dimensional vector field~\citep{dof}. It measures the spatial and temporal changes of intensity to look for a matching pixel in the next frame
with the assumption that all the temporal intensity changes in the image are due to motion only. In general, there are three common assumptions to approximate optical flow values: (1) Brightness constancy -- the observed brightness of the objects are constant over time (shadow and illuminations due to any motion are constant);  (2) Spatial coherence -- the surfaces of an object have spatial extent where its neighboring points are likely to be of the same surface hence having similar velocity values; and (3) Temporal persistence -- image motion of a surface patch changes gradually through time. 

\cite{pof} did the performance comparison among the four basic optical flow techniques, namely: region-based matching, differential, energy-based and phase-based method. Local differential method was found to be the most reliable and produced consistent results. Therefore, we opt for the \emph{differential method} \citep{bainbridge1997determine} to approximate the optical flow motion vectors. The general optical flow constraint equation is defined as: 

\begin{equation} \label{eq:flow}
\nabla I \bullet \vec{p} + I_{t} =0,
\end{equation}

\noindent
where $I(x,y,t)$ is the image intensity at time $t$ 
located at spatial point $(x,y)$. $\nabla I$ = {\it($I_{x},I_{y}$)} is the spatial gradients and {\it $I_{t}$} is the temporal gradient of the intensity function. Assume that the point of interest in the image is initially positioned at $(x,y)$, and it moves through a distance $(dx,dy)$ after the change in time of $dt$. The flow vector $\vec{p}$ consists of its horizontal and vertical components, $\vec{p} = [p = dx/dt,q = dy/dt$]\textsuperscript{\it T} denotes the horizontal and vertical components of the optical flow.

\subsection{Optical Strain}
Optical strain is a technique derived from the concept of optical flow that enables the computation of small and subtle motion on the face by measuring the amount of facial tissue deformation. 
This is in part motivated by the concept of strain rate tensor~\citep{ogden1997} from continuum mechanics, which describes the rate of change of the deformation of a material in the vicinity of a certain point, at a certain moment of time.
The work by~\cite{towards} demonstrated that optical strain is more reliable than optical flow in the task of automatic macro and micro facial expression spotting, producing more consistent results in their experiments. In this work, we intend to leverage the strengths of optical strain to describe suitable features for detection and recognition tasks.

Optical strain can be described by a two-dimensional displacement vector, ${\bf u} = [u,v]\textsuperscript{\it T}$. The magnitude of the optical strain can be represented in the form of a \emph{Lagrangian strain tensor} \citep{ogden1997}:

\begin{equation} \label{eq:tensor}
\varepsilon = \frac{1}{2} [\nabla \bf u + (\nabla \bf u)^{\it T} ]
\end{equation}
which in expanded form, is defined as,
\begin{equation}
\varepsilon = \begin{bmatrix}
      		\varepsilon_{xx} = \frac{\partial u}{\partial x} & \varepsilon_{xy} = \frac{1}{2}(\frac{\partial u}{\partial y} + \frac{\partial v}{\partial x}) \\[1em]
      	    \varepsilon_{yx} = \frac{1}{2}(\frac{\partial v}{\partial x} + \frac{\partial u}{\partial y}) & \varepsilon_{yy} = \frac{\partial v}{\partial y}
     		\end{bmatrix}
\end{equation}

\noindent
where $\{\varepsilon_{xx},\varepsilon_{yy}\}$ are the normal gradient components while $\{\varepsilon_{xy},\varepsilon_{yx}\}$ are the shear components of the optical strain.


Each of these strain components are a function of the displacement vectors ({\it u,v}). Thus, they can be approximated using the flow vector components ({\it p,q}) from Eq.~(\ref{eq:flow}) in discrete form, where $\Delta t$ is the time instance between two image frames: 

\begin{equation}\label{eq:p}
p = \frac{dx}{dt} \doteq \frac{\Delta x}{\Delta t} = \frac{u}{\Delta t} , u = p \Delta t,
\end{equation}

\begin{equation} \label{eq:q}
q = \frac{dy}{dt} \doteq \frac{\Delta y}{\Delta t} = \frac{v}{\Delta t} , v = q \Delta t
\end{equation}

\noindent

By setting $\Delta t$ to a constant interval length, the partial derivatives of Eq.~(\ref{eq:p}) and~(\ref{eq:q}) can be approximated, respectively as:

\begin{equation}
\frac{\partial u}{\partial x} = \frac{\partial p}{\partial x}\Delta t, \frac{\partial u}{\partial y} = \frac{\partial p}{\partial y}\Delta t,
\end{equation}

\begin{equation}
\frac{\partial v}{\partial x} = \frac{\partial q}{\partial x}\Delta t, \frac{\partial v}{\partial y} = \frac{\partial q}{\partial y}\Delta t
\end{equation}

The second order derivatives are determined using finite difference approximations, i.e.

\begin{equation}
\begin{split}
\frac{\partial u}{\partial x} &= \frac{u(x+\Delta x)-u(x-\Delta x)}{2 \Delta x} \\
&\doteq \frac{p(x+\Delta x)-p(x-\Delta x)}{2 \Delta x}
\end{split}
\end{equation}

\begin{equation}
\begin{split}
\frac{\partial v}{\partial y} &= \frac{v(y+\Delta y)-v(y-\Delta y)}{2 \Delta y} \\
&\doteq \frac{q(y+\Delta y)-q(y-\Delta y)}{2 \Delta y}
\end{split}
\end{equation}

\noindent
where $\Delta x = \Delta y =  1$ pixel. The partial derivatives $\partial u/\partial y$ and $\partial v/\partial x$ are calculated in the similar manner.

Finally, the optical strain magnitude for each pixel at a particular time $t$ can be calculated as follows:

\begin{equation}
\varepsilon = \sqrt{{\varepsilon_{xx}}^{2} + {\varepsilon_{yy}}^{2} + {\varepsilon_{xy}}^{2} +{\varepsilon_{yx}}^{2}}.
\label{eq:osm}
\end{equation}

\begin{figure}[t!]
\centering
\includegraphics[width=65mm]{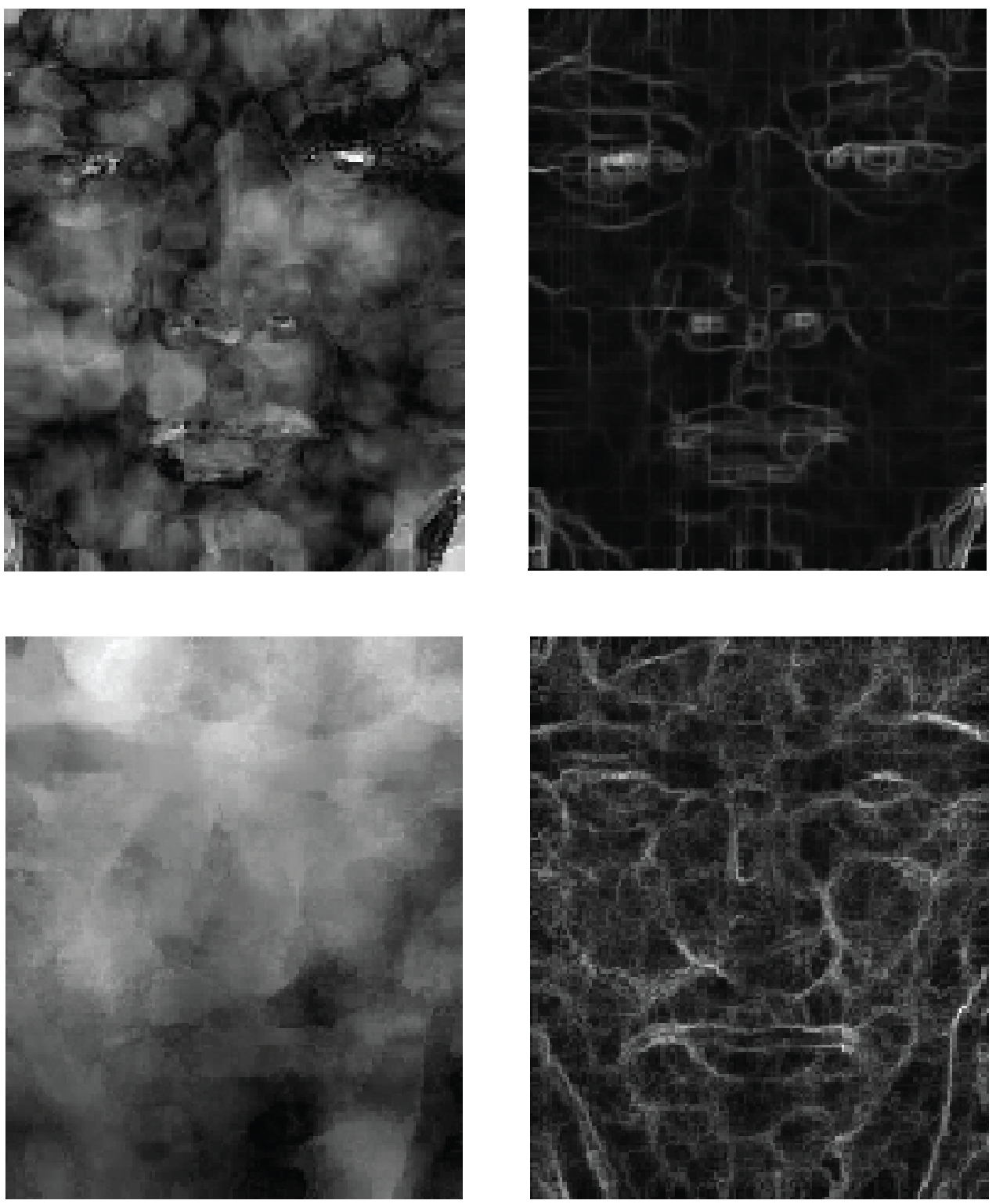}
\caption{Example of optical flow (left) and optical strain (right) images from CASME (top row) and SMIC (bottom row) databases.}
\label{fig:os}
\end{figure}

\noindent Fig. \ref{fig:os} shows the optical flow and optical strain images obtained from two sample videos. The values from both images are temporally sum-pooled and normalized for better visualization. In the bottom row images, the raised left eyebrow that is somewhat indistinguishable with the optical flow fields, is clearly emphasized by the strain components.

\subsection{Block-based LBP-TOP}
Local facial features on a video or sequence of images can be represented using the state-of-the-art dynamic texture descriptor - Local Binary Patterns from Three Orthogonal Planes (LBP-TOP)~\citep{dynamic}. The LBP code is extracted from three orthogonal planes (i.e., XY, XT and YT), which encodes the appearance and motion in three directions. Each pixel in an image forms a LBP code by applying a thresholding technique within its neighbourhood:  
\begin{equation}
\label{eq:lbp}
LBP_{P,R} = \sum\limits_{p=0}^{P-1} s(g_p - g_c)2^p , s(x) = \begin{dcases*}
1,  & x $\geq$ 0\\
0, & x $<$ 0
\end{dcases*}
\end{equation}

\noindent 
where $P$ is the number of neighbouring points around the center pixel, $(P,R)$ representing a neighbourhood of $P$ points equally spaced on a circle of radius $R$, $g_c$ is the gray value of the center pixel and $g_p$ are the $P$ gray values of the sampled points in the neighbourhood.

\begin{figure*}[t!]
\centering
\includegraphics[width=150mm]{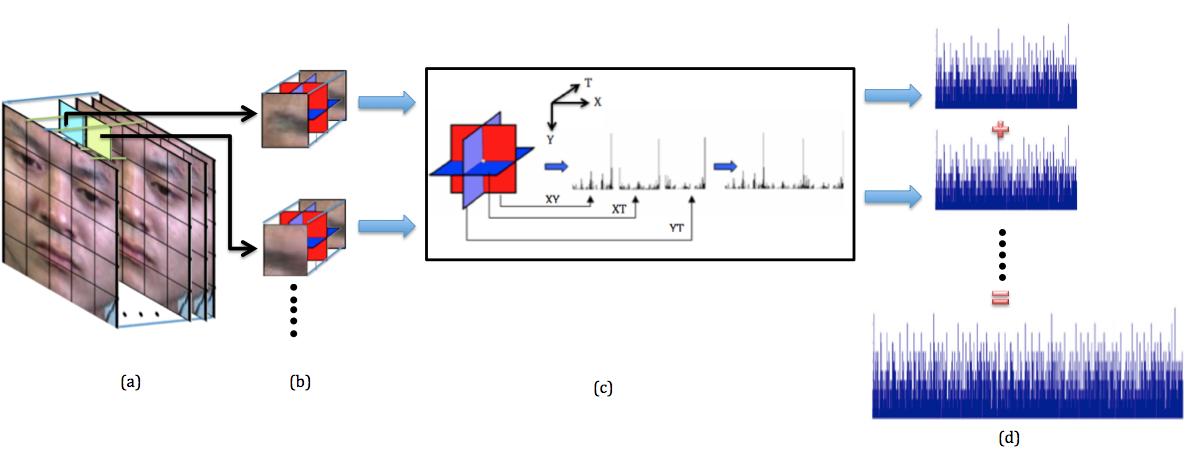}
\caption{Block-based features extraction of the first two block volumes in three dimension space of a video sequence: (a) Block volumes; (b) LBP features from three orthogonal planes; (c) Concatenation of the histograms from XY, XT and YT planes to become a single histogram; (d) Concatenation of the feature histograms from the two block volumes}
\label{fig:LBPTOP}
\end{figure*}

Block-based LBP-TOP partitions each image appearance plane (XY) into $N\times N$ non-overlapping blocks then concatenates the feature histograms of each volumetric blocks to construct the final resultant histogram. This is done for each of the three orthogonal planes. Fig.~\ref{fig:LBPTOP} illustrates the process of extracting and concatenating the computed local features from the first two blocks in three orthogonal planes. The concatenated histogram describing the global motion of the face over a video sequence can be succinctly denoted as,  

\begin{equation}
\begin{split}
M_{b_{1},b_{2},d,c} = \sum\limits_{x,y,t}I\{{h_{d}(x,y,t)=c}\}, \\
c=0, \dots,2^P-1; \;d=0,1,2; \; b_{1},b_{2}\in1 \ldots N
\end{split}
\end{equation}

\noindent
where $2^P$ is the number of different labels produced by the LBP operator on the $d$-th plane ($d=0:\text{XY}$ (appearance)$, 1: \text{XT}$ (horizontal motion) and $2:\text{YT}$ (vertical motion)). As the LBP code cannot be computed at the borders of the 3-D video volume, only the central part is taken into consideration. $h_{d}(x,y,t)$ is the LBP code, i.e., Eq. (\ref{eq:lbp}), of the central pixel $(x,y,t)$ on the $d$-th plane, {$x \in \{0,\dots,X-1 \}, y \in \{0,\dots,Y-1\} ,t \in \{0,\dots,T-1\}$}. $X$ and $Y$ are the width and height of image (thus, $b_{1}$ and $b_{2}$ are the row and column index, respectively), while $T$ is the video length. As such, the functional term $I\{A\}$ determines the count of the $c$-th histogram bin when $h_{d}(x,y,t)=c$:
\begin{equation}
I\{A\} = \begin{dcases*}
        1,  & if $A$ is true\\
        0, & if $A$ is false
        \end{dcases*}
\end{equation}
Hence, the final feature histogram is of dimension $2^P\times 3N^2$. For an appropriate comparison of the features among video samples of different spatial and temporal lengths, the concatenated histogram is sum-normalized to obtain a coherent description:

\begin{equation}
\overline{M}_{b_{1},b_{2},d,c} = \frac{M_{b_{1},b_{2},d,c}}{\sum\limits_{k=0}^{n_{d}-1}M_{b_{1},b_{2},d,k}}
\label{eq:blockhist}
\end{equation}

Throughout this paper, we will denote the LBP-TOP parameters by $\text{\it LBP-TOP}_{P_{XY},P_{XY},P_{YT},R_X,R_Y,R_T}$ where the $P$ parameters indicate the number of neighbour points for each of the three orthogonal planes, while the  $R$ parameters denote the radii along the X, Y, and T dimensions of the descriptor.

\section{Proposed Algorithm}
\label{sec:proposed}
We propose three main steps to extract the spatio-temporal features by utilizing facial optical strain information: (1) The optical strain magnitudes in each frame are derived from the optical flow values. Then all the optical strain maps in each video are temporally pooled
into a composite strain map. Thereafter, the optical strain magnitudes in the composite strain map are directly used as features; (2) Spatio-temporal pooling is applied on the optical strain frames of each video, then the final matrix of normalized coefficient values obtained are used as the weights for each video. The weighting matrix (of $N\times N$ dimension after pooling) is then multiplied with their respective LBP-TOP-extracted histogram bins on the XY plane; (3) Finally, the feature histograms extracted in steps (1) and (2) are concatenated into a final feature histogram that represents the processed video sample. 

\begin{figure*}[t!]
\centering
\includegraphics[width=150mm]{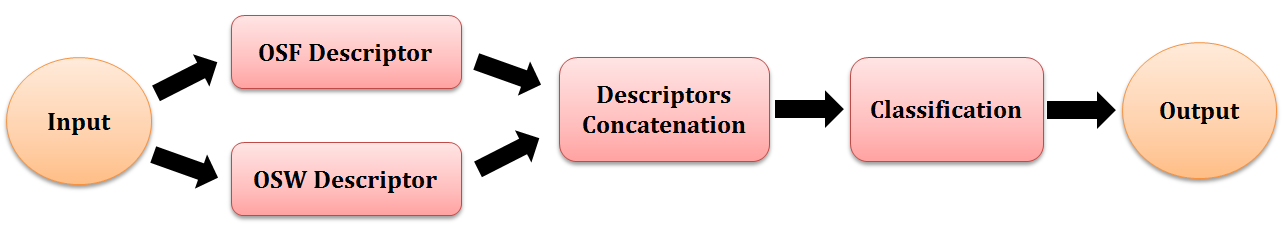}
\caption{Overview of the proposed algorithm}
\label{fig:block_diagram}
\end{figure*}

The overview of the proposed method is shown in Fig.~\ref{fig:block_diagram}. Note that, all the image data from the prior-database \citep{casme2,smic} are captured under constrained lab condition and have been performed face registration and alignment.

\subsection{Optical Strain Features (OSF)} 
\label{sec:osf}

As optical strain magnitudes can aptly describe the minute extent of facial deformation at the pixel level, they can directly be employed as features as well. For clarity, we first describe the notations used in the subsequent sections.

A micro-expression video clip is expressed as 
\begin{equation}
s_{i} = \{f_{i,j} | i=1,\dots,n; j=1,\dots ,F_{i}\}
\end{equation}
where $F_{i}$ is the total number of frames in the ${i}$-th sequence, which is taken from a collection of $n$ video sequences. 

The optical flow field is first estimated by its 2-D motion vector, $\vec{p}=(p,q)$. Then, the optical strain magnitude at each pixel location $\varepsilon_{x,y}$ (from Eq.~(\ref{eq:osm})) is calculated for each flow field over two consecutive frames, i.e., $\{f_{i,j},f_{i,j+1}\}$. 

Hence, each video of resolution $X\times Y$ produces a set of $F_{i}-1$ optical strain maps, each denoted as  
\begin{equation}
m_{i,j} = \{\varepsilon_{x,y} | x=1,\dots,X; y=1,\dots ,Y\}
\end{equation}
for $i \in 1,\dots ,n$ and $j \in 1,\dots ,F_{i}-1$

The following steps first describe essential pre-processing steps for noise reduction and signal attenuation, followed by how optical strain features (OSF) are obtained.

\subsubsection{Pre-processing}
Prior to feature extraction, two pre-processing steps are carried out to reduce unwanted noise and
attenuate the strong and weak signals from the optical strain maps.

First, the edges in each 
strain map $m_{i,j}$ are removed. The edges are the gradient of the moving objects that consist of local maximas. Hence, the purpose of eliminating the edges is to remove a large number of irrelevant maximas if the strain map is very noisy~\citep{barcelos2003well}.
Among the different types of edge detectors, the Sobel filter justifies its feasibility by two main advantages~\citep{sobel}--- its ability to detect the edges in a noisy image by introducing smoothing and blurring effect on the image; the differential of two rows or two columns enhances the strength of important edges. The Sobel operator is a simple  approximation to the concept of 2-D spatial gradient, by convoluting a grayscale input image with a pair of 3$\times$3 convolution masks~\citep{sobel2}. 
Inspired by \cite{happy2015automatic}, they demonstrated that horizontal edge detector always generates a distinct edge on macro facial expression databases (i.e., CK+ \citep{lucey2010ck} and JAFFE \citep{lyons1998jaffe}). Besides, we conducted experiments on the micro-expression databases to compare the performance of horizontal and vertical edges. We empirically discovered that removing the vertical edges generated better recognition results than removing the horizontal edges only as well as both the horizontal and vertical edges. Therefore, vertical direction derivatives of Sobel edge detector was used in our experiments.

Secondly, 
the magnitudes in each optical strain map $m_{i,j}$ are clipped to zero for $\varepsilon_{x,y} \notin [T_l, T_u]$, with the two threshold values $T_l$ and $T_u$ denoting the lower and upper thresholds respectively. 

\begin{figure}[t!]
\centering
\includegraphics[width=80mm]{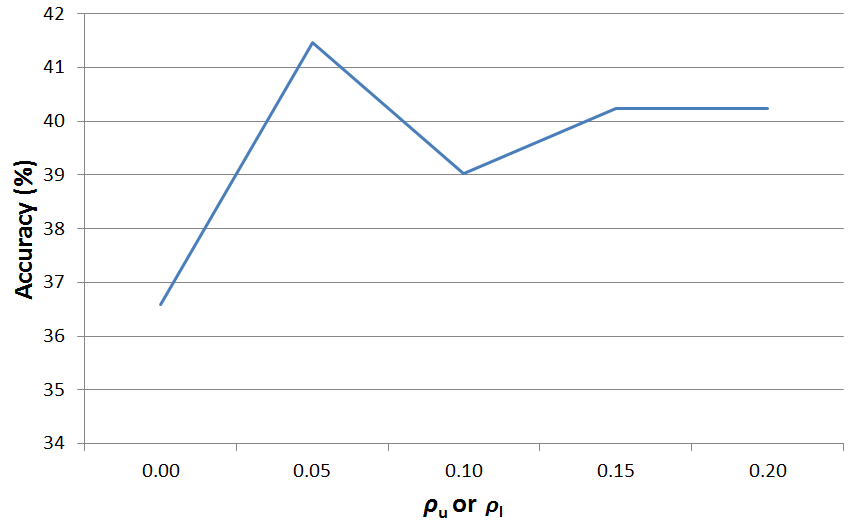}
\caption{Effect of $\rho_{l}$ and $\rho_{u}$ values on micro-expression recognition rate for the SMIC database}
\label{fig:threshold}
\end{figure}

The values of $T_l$ and $T_u$ are determined using the lower and upper percentages $(\rho_l,\rho_u)$ 
of the strain magnitude range, $[\varepsilon_{min} = \text{min}\{\varepsilon_{x,y}\} , \varepsilon_{max} = \text{max}\{\varepsilon_{x,y}\}]$
The lower and upper thresholds are computed as follows:
\begin{equation}\label{eq:threshold}
T_{l} = \varepsilon_{min} + \rho_{l} \cdot (\varepsilon_{max} - \varepsilon_{min}), \rho_{l} \in[0,1]
\end{equation}
\begin{equation}
T_{u} = \varepsilon_{max} - \rho_{u} \cdot (\varepsilon_{max} - \varepsilon_{min}), \rho_{u} \in[0,1]
\label{eq:upper-threshold}
\end{equation}
Figure \ref{fig:threshold} illustrates the effect of $\rho_{l}$ and $\rho_{u}$ on the micro-expression recognition rate. It is observed that $\rho_{l}=\rho_{u}=0.05$ yields the best results. Therefore, we set the clipping tolerance to $5\%$ of the magnitude range of each processed frame.

\begin{figure}[t!]
\centering
\includegraphics[width=25mm]{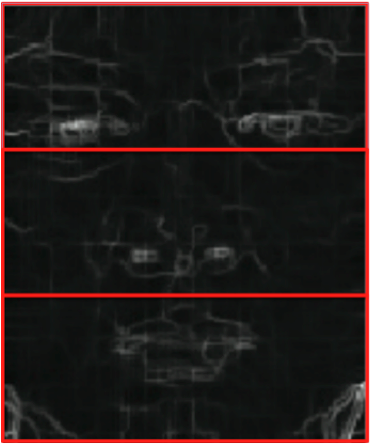}
\caption{Example of vertical segmentation of the optical strain frame into three regions.}
\label{fig:segment}
\end{figure}

With each frame properly aligned, the optical strain maps can be then segmented vertically into three regions of equal size (i.e. forehead--lower eyelid, lower eyelid--nostril and nostril--mouth) to obtain their individual local threshold values. 
The purpose of performing this segmentation is to minimize the effects of dominant motions that arise from a particular region as the range of strain magnitudes differ across the three regions. Fig.~\ref{fig:segment} shows how an optical strain map is divided into the three vertical segments. 


\begin{figure*}[t!]
\centering
\includegraphics[width=150mm]{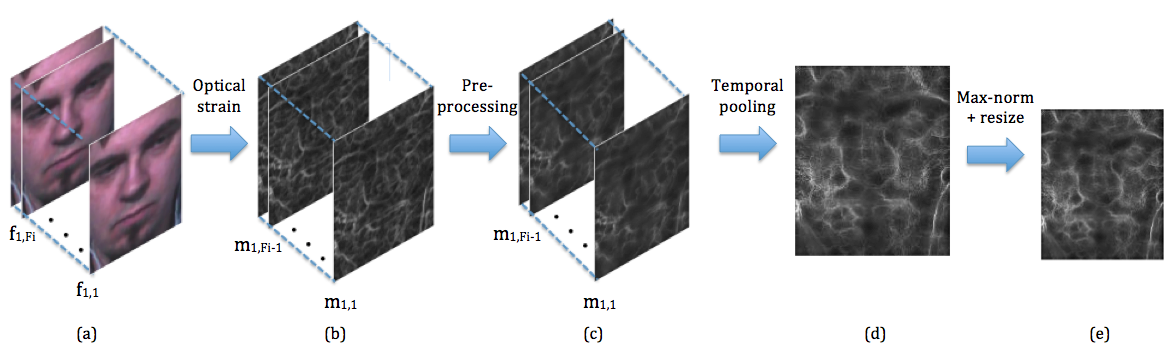}
\caption{Extracting optical strain features (OSF) from a sample video sequence: (a) Original images, $f_{1,j}$; (b) Optical strain maps $m_{1,j}$; (c) Images after pre-processing; (d) Temporal pooled strain images; (e) After applying maximum normalization and resizing}
\label{fig:OSF}
\end{figure*}

\subsubsection{Extracting Optical Strain Features}

In order to describe the optical strain patterns in a compact and consistent representation, the optical strain maps $m_{i,j}$ are pooled across time (i.e. temporal pooling).We perform temporal mean pooling to obtain a composite strain map,
\begin{equation}
\hat{m}_{i} 
= \frac{1}{F_{i}-1}\sum\limits_{j=1}^{F_{i}-1} 
m_{i,j}
\end{equation}
where all optical strain magnitudes $\varepsilon_{x,y}$ for each strain map $m_{i,j}$ are averaged across the temporal dimension. \js{The intuition behind this pooling step is to help in accentuating the minute motions in micro-expressions by aggregation of these facial strain patterns.} Mean pooling also ensures that the optical strain magnitudes are normalized based on their respective sequence lengths. 
Then, the composite strain map is max-normalized to increase the significance of its values. In the final step, we resize the composite strain map to 50x50 pixels and vectorize its rows to form a 2500-dimension feature vector. Fig. \ref{fig:OSF} shows a graphical illustration of the entire process of extracting optical strain features.

\subsection{Optical Strain Weighted (OSW) Features}

\label{sec:osw}


%

While the OSF describes pixel-level motion features, the LBP-TOP is capable of encoding texture dynamics in larger facial patches. In block-based LBP-TOP \citep{dynamic}, the feature histograms obtained from all blocks are given equal treatment. Since subtle \christy{expression}s typically occur in highly localized regions of the face (and this differs for different \christy{expressions} classes), the feature histogram representing these regions should be amplified. As such, larger motions will generate larger optical strain magnitudes and vice versa. A set of weights are then computed to scale the features in each block proportionally to their respective motion strengths. We proceed to elaborate how optical strain weighted (OSW) features are obtained.

\subsubsection{Extracting Block-based LBP-TOP Features}
Features are extracted by block-based LBP-TOP from each video clip $s_{i}$,
whereby the entire video volume is partitioned into $N \times N$ non-overlapping block volumes. For each of these block volumes, we compute the LBP features from three orthogonal planes (concatenated to form LBP-TOP) to obtain dynamic texture features that are local to each particular block region. Finally, the feature histograms from all $N\times N$ block volumes are concatenated to form the final feature histogram. 

\subsubsection{Pre-processing}

\begin{figure}[t!]
\centering
\includegraphics[width=85mm]{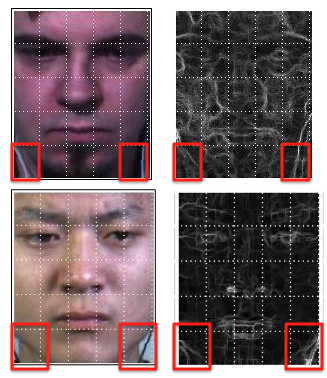}
\caption{Top row: a sample image from SMIC (left) and the corresponding optical strain map (right). Bottom row: a sample image from CASME II (left) and the corresponding optical strain map (right)}
\label{fig:noise}
\end{figure}

Upon partitioning into blocks, two blocks that are located at the left and right bottom corner of the frames are eliminated due to 
noticeable amount of movements or noise caused by the background lighting condition, and also the presence of wires from the headset worn by the subjects (see Fig.~\ref{fig:noise} for a frame sample from both datasets). For ease, we refer to these two blocks as ``noise blocks". Therefore, we omit these noise blocks and only the remaining $N^2-2$ blocks are utilized towards building the feature histogram.  

All the images are filtered by a Gaussian filter to reduce pixel noise. The filter size applied was $5\times 5$ pixels and standard deviation $\sigma=0.5$ in order to reduce the existing background noises prior to processing. This is because the motions characterized by the subtle facial expressions are very fine, it is likely that these noises might be incorrectly identified as fine facial motions.

\subsubsection{Determining Weights by Spatio-temporal Pooling}
To obtain the weights for each block, we perform spatio-temporal pooling on all optical strain maps $m_{i,j}$ in the video sequence. We consider spatio-temporal pooling in a separable fashion; spatial mean pooling is performed first, followed by temporal mean pooling.

Firstly, spatial mean pooling averages all the strain magnitudes $\varepsilon_{x,y}$ within each block, resulting in a block-wise strain magnitude:
\begin{equation}
z_{b_{1},b_{2}} = \frac{1}{HL}\sum\limits_{y=(b_{2}-1)H+1}^{b_{2}H}\; \sum\limits_{x=(b_{1}-1)L+1}^{b_{1}L} \varepsilon_{x,y}
\end{equation}
\noindent where $L=\frac{X}{N}$, $H=\frac{Y}{N}$, the block indices $(b_{1},b_{2})\in 1 \ldots N$, and $(X,Y)$ are the dimensions (width and height) of the frame. This process summarizes the encoded features locally in each block area of the face.  

\begin{figure*}[t!]
\centering
\includegraphics[width=150mm]{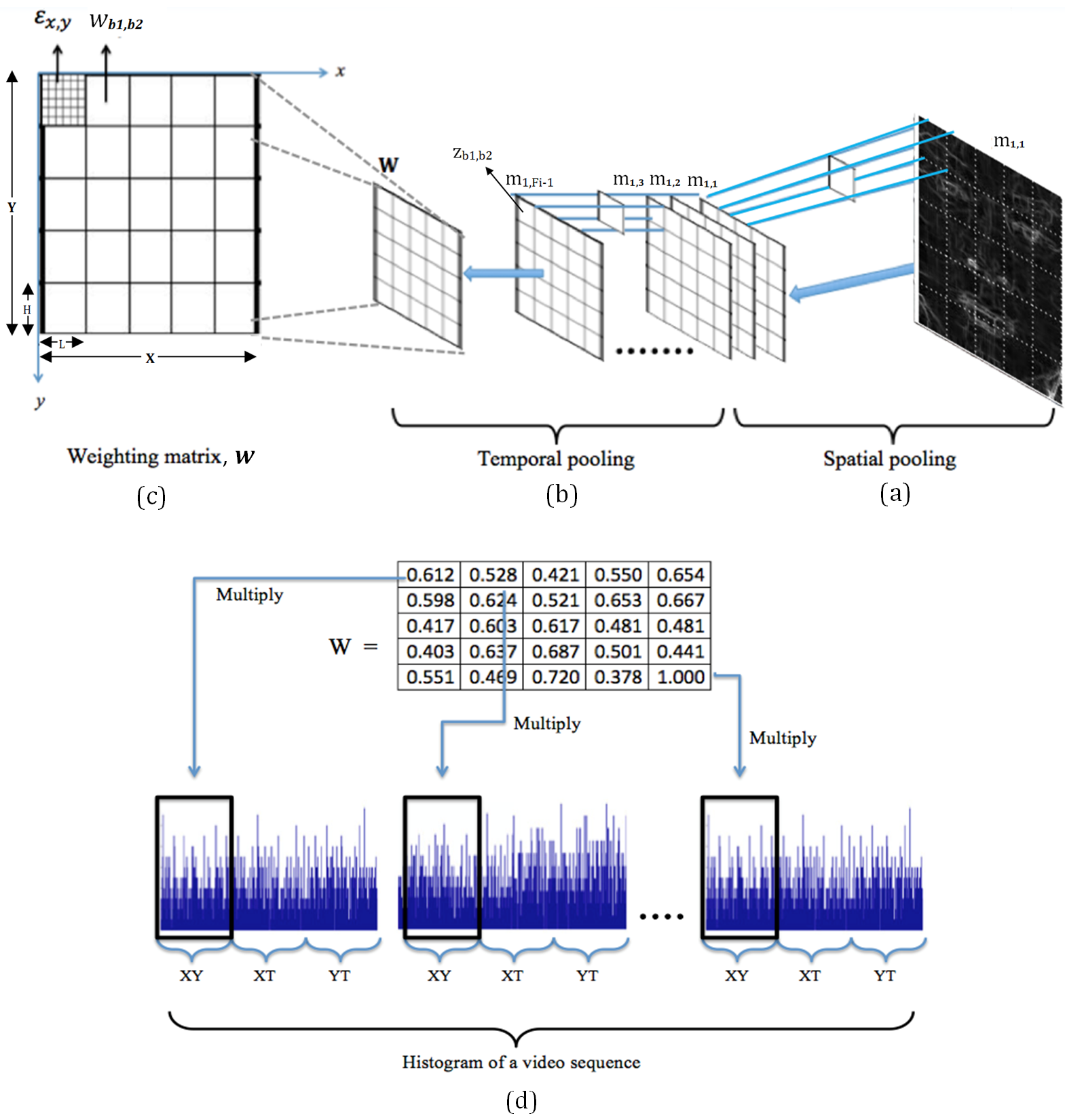}
\caption{\christy{Derivation of the optical strain weighted feature for the first video}: (a) Each $j$-th frame in $m_{1,j}$ is divided into $5 \times 5$ non-overlapping blocks then the values of $\varepsilon_{x,y}$ within each block region are spatially pooled; (b) The block-wise strain magnitudes $z_{b_1,b_2}$ from all frames $(j\in 1\dots F_{i-1})$ are temporally mean pooled; (c) The $N\times N$ size of weighting matrix, $\bf W$ is formed; (d) coefficients of $\bf W$ are multiplied to their respective $XY$-plane histogram bins.}
\label{fig:OSW}
\end{figure*}

Secondly, temporal mean pooling is applied on the spatially-pooled frames, where the values of $z_{b_{1},b_{2}}$ are averaged up along the temporal dimension across all video frames. Therefore, for each video, we derive a unique set of $N\times N$ weights, $\textbf{W}_{i}=\{w_{b_{1},b_{2}}\}_{b_{1},b_{2}=1}^{N}$ where each weight coefficient $w$ is described as:
\begin{align}
w_{b_{1},b_{2}} &= \frac{1}{F_{i}-1}\sum\limits_{t=1}^{F_{i}-1}  z_{b_{1},b_{2}} \nonumber \\
&= \frac{1}{(F_{i}-1)HL}\sum\limits_{t=1}^{F_{i}-1} \sum\limits_{y=(b_{2}-1)H+1}^{b_{2}H}\; \sum\limits_{x=(b_{1}-1)L+1}^{b_{1}L} \varepsilon_{x,y}
\end{align}

\subsubsection{Weighted \texorpdfstring{$XY$-Plane Histogram}{Lg}}
After obtaining the feature histograms extracted by LBP-TOP and the optical strain weights, the coefficients of the weight matrix $\mathbf{W}$ are multiplied with the XY-plane feature histograms of their corresponding matching blocks.
This weighting procedure is performed only on features from the XY plane so that the motion strengths are well accentuated in each local area of the face.

Concisely, the optical strain weighted histograms can be defined as:
\begin{equation}
G_{b_{1},b_{2},d,c} = \begin{dcases*}
        w_{b_{1},b_{2}}\overline{M}_{b_{1},b_{2},d,c},  & if $d$ = 0;\\
        \overline{M}_{b_{1},b_{2},d,c}, & otherwise.
        \end{dcases*}
\end{equation}
where $\overline{M}$ is the normalized feature histogram of block $(b_1,b_2)$ from Equation \ref{eq:blockhist}. The whole process flow of obtaining the optical strain weighted features is graphically shown in Fig.~\ref{fig:OSW}.

\subsection{Concatenating OSF and OSW Features
}
In the final step, the two extracted features --- OSF and OSW features are concatenated into a single composite feature histogram. The concatenation process enriches the variety of features used, providing further robustness towards the detection and recognition of facial micro-expressions. The dimension of the feature histogram in LBP-TOP with 5$\times$5 block partitions are $5\times 5 \times 3\times 15 \text{(OSW)} + 50\times50 \text{(OSF)}=3625$.


\section{Experiment}
\label{sec:experiment}
\subsection{Datasets}
The experiments were performed on two spontaneous micro-expressions datasets that are publicly available--- CASME II~\citep{casme2} and SMIC~\citep{smic}. 

\subsubsection{CASME II}
The Chinese Academy of Sciences Micro-Expression (CASME) II dataset contains 247 micro-expressions film clips elicited from 26 participants (mean age of 22.03 years, standard deviation of 1.60). Each clip only consists of one type of \christy{expression} and they were recorded using a Point Grey GRAS-03K2C camera with a high temporal resolution of 200 \emph{fps} and a spatial resolution of 640$\times$480 pixels. To collect the micro-expression videos, some of the participants were asked to keep neutralize faces and the rest were asked to suppress their facial motion when watching the video films. The clips are labeled with five \christy{expression} classes: happiness, disgust, surprise, repression and tense. The ground truths of the dataset are provided, which include the onset, offset of the \christy{expression}, the represented \christy{expression} and the marked action unit (AUs). The labeling of the micro-expressions were done by two coders. They determined the micro-expressions from the raw data by first spotting the onset and offset frames for each video. Then, the spotted sequences are labeled as micro-expressions if the duration lasts for less than 1 \emph{s}. The decision of marking them as micro-expressions as well as identifying their respective classes are based on the AUs found, the participants' self-report and the contents of the film clips. 

A five-class baseline recognition accuracy was reported as 63.41\%. Block-based LBP-TOP was employed as the feature extractor with images partitioned into 5 $\times$ 5 blocks. The classifier used for baseline evaluation is Support Vector Machine (SVM) with leave-one-video-out cross-validation (LOVOCV) setting.

\subsubsection{SMIC}
\label{sec:smic}
The second dataset called Spontaneous Micro-expression (SMIC) Database consists of 16 participants (mean age of 28.1 years, six females and ten males with eight Caucasians and eight Asians) and 164 micro-expressions videos. The video clips were recorded using a high speed PixeLINK PL-B774U camera at a reasonably good resolution of 640$\times$480 and frame rate of 100 \emph{fps}. The micro-expressions were collected by asking the participants to put on a poker face and suppress their genuine feelings when watching the films. The video clips are classified into three main categories: positive (happiness), negative (sad, fear, disgust) and surprise. Labeling of the micro-expression clips was done by two coders, whereby the clips were first viewed frame by frame at a slower playback before repeating it in increasing frame rate. This follows the suggestion established by~\cite{ekman}. Next, the selected clips agreed by both of the coders were compared to the participants' self-reported \christy{expression}s before deciding to include the micro-expression clips into the database. In the pre-processing step, all the video clips were first interpolated into ten frames using the temporal interpolation model (TIM)~\citep{smic} to minimize feature complexity and processing time. In addition, TIM was proved to boost the micro-expression recognition performance~\citep{york}. \christy{Hence, we also apply TIM to standardize each video in the SMIC database to a length of 10 frames.} the Two tasks were performed on the database -- detection and recognition, where the block-based LBP-TOP ($N\times N$ non-overlapping blocks) is chosen for feature extraction and SVM as classifier. Experiments were conducted based on leave-one-subject-out cross-validation (LOSOCV) setting.

The best baseline performance for this three-class classification task is 48.78\%. An $8\times 8$ block partitioning was used to obtain the best results. On the other hand, the presence of micro-expressions (i.e., detection task) can also be distinguished by comparing against a supplementary dataset that contains non-micro-expressions. The non-micro-expression clips have the same length and distribution as the micro-expressions clips and it includes several types of movements such as general facial expressions, eye blinking, lip sipping and slight head tilting. The baseline detection rate reported in the paper is 65.55\%, based on a $5\times 5$ block partitioning. 

\section{Results and Discussion}
\label{sec:results}
We evaluated our methods on two separate experiments: (1) detection of micro-expressions (SMIC only), and (2) recognition of micro-expressions (CASME II and SMIC). The detection task involves determining
whether the expression shown in a clip contains micro-expressions or not, regardless of the emotional state it represents. Meanwhile, the recognition task involves identification of the emotional state present in the video clip. 

\christy{\textcolor{black}{Note that both CASME II and SMIC databases provide the cropped face video sequence, where only the face region is retained while the unnecessary background are removed.} We directly use the cropped image frames in our experiments; these frames have an average spatial resolution of $340 \times 280$ for CASME II and $170 \times 140$ pixels for SMIC. Here, we establish the parameter setting used in our experiments, namely, the parameters used in the feature extractor and classifier are mostly the values adopted from the original work, i.e., CASME II~\citep{casme2} and SMIC~\citep{smic}. SVM with linear kernel ($c=10000$) was utilized as classifier. The block sizes of LBP-TOP were selected based on the original works~\citep{casme2,smic}. We set the block partitioning of LBP-TOP to both $5 \times 5$ and $8 \times 8$. In addition, the number of neighbour points and the radii along the three orthogonal planes were set to $\text{\it LBP-TOP}_{4,4,4,1,1,4}$. The reason of selecting these parameter configurations is explained in Section~\ref{sec:discussion}.}

In SMIC, recognition is a three-class classification (i.e., positive, negative, surprise classes), while detection of micro-expressions is a binary decision (i.e., yes/no). Evaluations on SMIC were conducted using SVM classifier with LOSOCV setting. For CASME II, only the recognition task was performed since the database did not provide non-micro-expression clips to enable a detection task. The recognition task on CASME II is a five-class problem, evaluated using SVM classifier with LOVOCV setting. 

There are two ways to measure the classification performance in the LOSOCV setting, namely, \emph{macro}- and \emph{micro}-averaging \citep{tsoumakas2010mining}. The macro-averaged result gives the accuracy computed by averaging across all individual subject-wise accuracy results. Micro-averaged result refers to the overall accuracy result across all evaluated samples. We also present further performance metrics such as F1-measure, precision and recall when the LOSOCV setting is used, as suggested by \cite{le2014spontaneous}. These metrics provide a more meaningful perspective than accuracy rates when the datasets used are naturally imbalanced since each subject has a different number of video samples.

The three proposed methods: (i) Optical Strain Features ($\bf OSF$), (ii) Optical Strain Weighted ($\bf OSW$) LBP-TOP Features, and (iii) concatenation of both features ($\bf {OSF + OSW}$), were evaluated and compared to their respective baseline methods on both detection and recognition experiments.


\subsection{Detection and Recognition Results}


\setlength{\tabcolsep}{4pt}
\begin{table}[tb]
\begin{center}
\caption{Micro-expression detection and recognition results on SMIC and CASME II database with LBP-TOP of $5\times 5$ block partitioning}
\label{table:5x5}
\begin{tabular}{lccccc}
\noalign{\smallskip}
\hline
\multirow{2}{*}{Methods} & \multicolumn{2}{c}{Det - SMIC} & \multicolumn{2}{c}{Recog - SMIC} & Recog \\ 
\cline{2-5}
& Micro & Macro & Micro & Macro & CASME II \\
\hline
\noalign{\smallskip}

Baseline  
&	60.67	&	66.23
&	40.85	&	42.85
&	61.94\\

OSF
&	66.16	&	66.34
&	41.46	&	46.00
&	51.01\\ 

OSW 
&	63.72	&	67.67
&	46.95	&	49.42
&	62.75\\ 

OSF+OSW 
&\bf71.95	&\bf74.52
&\bf43.90	&\bf48.09
&\bf63.16\\

\hline
\end{tabular}
\end{center}
\end{table}
\setlength{\tabcolsep}{1.4pt}
\setlength{\tabcolsep}{4pt}

\setlength{\tabcolsep}{4pt}
\begin{table}[tb]
\begin{center}
\caption{Micro-expression detection and recognition results on SMIC and CASME II database with LBP-TOP of $8\times 8$ block partitioning}
\label{table:8x8}
\begin{tabular}{lccccc}
\noalign{\smallskip}
\hline
\multirow{2}{*}{Methods} & \multicolumn{2}{c}{Det - SMIC} & \multicolumn{2}{c}{Recog - SMIC} & Recog \\ 
\cline{2-5}
& Micro & Macro & Micro & Macro & CASME II\\
\hline
\noalign{\smallskip}

Baseline  
&	57.32	&	59.35
&	45.73	&	47.76
&	61.13\\

OSF
&	66.16	&	66.34
&	41.46	&	46.00
&	51.01\\ 

OSW 
&	62.80	&	63.37
&	49.39	&	50.71
&	61.94\\ 

OSF+OSW 
&\bf72.87	&\bf74.16
&\bf52.44	&\bf58.15
&\bf61.54\\

\hline
\end{tabular}
\end{center}
\end{table}
\setlength{\tabcolsep}{1.4pt}
\setlength{\tabcolsep}{4pt}

From the results shown in Table \ref{table:5x5} and \ref{table:8x8}, we observe that the $\bf OSF$ method is able to produce reasonably positive results compared to the baselines 
in some cases. However, better and more consistent results were obtained using $\bf OSW$ and $\bf OSF+OSW$ methods for both macro- and micro-averaging measures. 

For the detection task on SMIC database (using $5\times 5$ blocks in LBP-TOP), $\bf OSF+OSW$ outperformed the baseline by $11.28\%$ and $8.29\%$ in micro- and macro-averaged results respectively. In addition, we obtained an even larger improvement of $\approx 15\%$ when the block partition of $8\times 8$ is used. Confusion matrix is a typical measurement that demonstrates the 
recognition performance of each individual 
class.
Table~\ref{table:mat_smic_det} lists the confusion matrices of the detection results on SMIC database. Using the proposed method $\bf OSF+OSW$, the micro- and non micro-expressions can be better 
distinguished. 
It can be seen that for non micro-expression, there is a great increase in recognition rate of around 17\%.

\begin{table}[tb!]

	\begin{center}
    \caption{Confusion matrices of baseline and $\bf OSF+OSW$ methods for detection task on SMIC database with LBP-TOP of $5\times 5$ block partitioning (measured by recognition rate \%)}
    \label{table:mat_smic_det}
    \begin{subtable}[l]{.48\linewidth}
    \centering
        \caption{Baseline}
        \begin{tabular}{|lcc|}
        \cline{1-3} 
         & micro	 & n/micro \\
        \hline

        micro  
        &\bf64.63	&	35.37\\

        n/micro 
        &	43.29	&\bf56.71	\\

        \hline
        \end{tabular}
    \end{subtable}%
    \begin{subtable}{.48\linewidth}
    \centering  
        \caption{OSF+OSW}
        \begin{tabular}{|lcc|}
        \cline{1-3} 
        & micro	 & n/micro \\
        \hline

        micro  
        &\bf70.12	&	29.88\\

        n/micro
        &	26.22	&\bf73.78	\\

        \hline
        \end{tabular}
    \end{subtable} 
    \end{center}
\end{table}

In the recognition experiment on the SMIC database, we are able to achieve up to $\approx 5\%$ improvement over the baseline results on the concatenated $\bf OSF+OSW$ method using $5\times 5$ in LBP-TOP (Table \ref{table:5x5}). This method also registered a performance improvement of $\approx 10\%$ over the baseline results when $8\times 8$ block partition is used (Table \ref{table:8x8}). These results point towards a significant improvement in feature representation when optical strain information is well-utilized. It is worth noting that although the $\bf OSF$ method did not perform as well, its contribution towards the concatenated $\bf OSF+OSW$ should not be disregarded. The detailed confusion matrices for the recognition performance on SMIC database are shown in Table~\ref{table:mat_smic_recog}. It can be seen that `Negative' and `Surprise' expressions can be recognized with higher accuracy using $\bf OSF+OSW$ method, while the accuracy of the `Positive' expression remains unchanged at 49.02\%. 

For the recognition experiment on the CASME II dataset, we observe a better performance by the $\bf OSW$ method as compared to the baseline and other evaluated methods. There is a substantial improvement in $\bf OSF+OSW$ method for both the $5\times 5$ and $8\times 8$ block partitions in LBP-TOP. Table~\ref{table:mat_casme_recog} shows the confusion matrices for the recognition results on CASME II database. We can observe `Disgust', `Tense' and `Surprise' can be recognized with higher accuracy using $\bf OSF+OSW$ method, but `Happiness' and `Repression' have lower recognition rate compared to the baseline. However, the average recognition accuracy for $\bf OSF+OSW$ method is better than the baseline.
 
Other performance metrics (including F1-score, recall and precision) reported in Table~\ref{table:5x5_f_r_p} and Table~\ref{table:8x8_f_r_p} further substantiate the superiority of our proposed methods over the baseline method, especially for the $\bf OSW$ and $\bf OSF+OSW$ methods in the detection and recognition tasks respectively, in SMIC database. The performance of the concatenated $\bf OSF+OSW$ method in CASME II recognition appears to be 
slightly better than
that offered by the baseline.

\begin{table*}[tb]

	\begin{center}
    \caption{Confusion matrices of baseline and $\bf OSF+OSW$ methods for recognition task on SMIC database with LBP-TOP of $8\times 8$ block partitioning (measured by recognition rate \%)}
    \label{table:mat_smic_recog}
    \begin{subtable}{.4\linewidth}
      \centering
        \caption{Baseline}
        \begin{tabular}{lccc}
        \noalign{\smallskip}
        \cline{1-4} 
         & negative	 & positive	&	surprise \\
        \noalign{\smallskip}
        \hline
        \noalign{\smallskip}

        negative  
        &\bf42.86	&	35.71	&	21.43	\\

        positive 
        &	37.25	&\bf49.02	&	13.73	\\
        
        surprise 
        &	32.56	&	20.93	&\bf46.51	\\

        \hline
        \end{tabular}
    \end{subtable}%
    \begin{subtable}{.75\linewidth}
      \centering
        \caption{OSF+OSW}
        \begin{tabular}{lccc}
        \noalign{\smallskip}
        \cline{1-4} 
         & negative	 & positive	&	surprise \\
        \noalign{\smallskip}
        \hline
        \noalign{\smallskip}

        negative 
        &\bf45.71	&	35.71	&	18.57	\\

        positive 
        &	39.22	&\bf49.02	&	11.76	\\
        
        surprise 
        &	23.26	&	9.30	&\bf67.44	\\

        \hline
        \end{tabular}
    \end{subtable} 
    \end{center}
\end{table*}

\begin{table*}[tb]

	\begin{center}
    \caption{Confusion matrices of baseline and $\bf OSF+OSW$ methods for recognition task on CASME II database with LBP-TOP of $5\times 5$ block partitioning (measured by recognition rate \%)}
    \label{table:mat_casme_recog}
    \begin{subtable}{\textwidth}\centering
      \centering
        \caption{Baseline}     
        \begin{tabular}{lccccc}
        \noalign{\smallskip}
        \cline{1-6} 
         & disgust	 & happiness	&	tense	&	surprise	&	repression \\
        \noalign{\smallskip}
        \hline
        \noalign{\smallskip}

        disgust  
        &\bf55.00	&	5.00	&	30.00	&	8.33	&	1.67\\

        happiness 
        &	6.06	&\bf54.55	&	15.15	&	0	&	24.24	\\
        
        tense 
        &	22.55	&	4.90	&\bf72.55	&	0	&	0	\\
        
        surprise 
        &	24.00	&	8.00	&	20.00	&\bf44.00	&	4.00	\\
       
       	repression 
        &	3.70	&	22.22	&	11.11	&	0	&\bf62.96	\\
        \hline
        \end{tabular}
    \end{subtable}%
    \vskip 8pt
    \begin{subtable}{\textwidth}\centering
      \centering
        \caption{OSF+OSW}
        \begin{tabular}{lccccc}
        \noalign{\smallskip}
        \cline{1-6} 
        & disgust	 & happiness	&	tense	&	surprise	&	repression \\
        \noalign{\smallskip}
        \hline
        \noalign{\smallskip}

        disgust  
        &\bf68.33	&	6.67	&	18.33	&	6.67	&	0\\

        happiness 
        &	12.12	&\bf39.39	&	30.30	&	3.03	&	15.15	\\
        
        tense 
        &	16.67	&	6.86	&\bf73.53	&	1.96	&	0.98	\\
        
        surprise 
        &	20.00	&	8.00	&	20.00	&\bf48.00	&	4.00	\\
       
       	repression 
        &	11.11	&	22.22	&	11.11	&	0	&\bf55.56	\\
        \hline
        \end{tabular}
    \end{subtable} 
    \end{center}
\end{table*}

\setlength{\tabcolsep}{4pt}
\begin{table*}[!tb]
\begin{center}
\caption{F1-score, recall and precision scores for detection and recognition performance on SMIC and CASME II database with LBP-TOP of $5\times 5$ block partitioning}
\label{table:5x5_f_r_p}
 \begin{subtable}{\textwidth}\centering
 \caption{Detection - SMIC} 
    \begin{tabular}{lccccccccccccccc}
    \noalign{\smallskip}
    \hline
    & \multicolumn{3}{c} {Micro} & \multicolumn{3}{c} {Macro}  \\
    \hline
    Methods 
    & F1 & Rec & Prec 
    & F1 & Rec & Prec\\
    \hline
    \noalign{\smallskip}

    Baseline  
    &	0.6070	&	0.6067	&	0.6074
    &	0.6623	&	0.6623	&	0.6697\\ 

    OSF
    &	0.6621	&	0.6616	&	0.6626
    &	0.6729	&	0.6634	&	0.6906\\ 

    OSW 
    &	0.6372	&	0.6372	&	0.6372
    &	0.6938	&	0.6767	&	0.7133\\ 

    OSF + OSW 
    &\bf0.7197	&\bf0.7195	&\bf0.7198
    &\bf0.7652	&\bf0.7452	&\bf0.7876\\ 
    \hline
    \end{tabular}
    \end{subtable}

    \vskip 10pt
    \begin{subtable}{\textwidth}\centering
    \caption{Recognition - SMIC} 
    \begin{tabular}{lccccccccccccccc}
    \noalign{\smallskip}
    \hline
    & \multicolumn{3}{c} {Micro} & \multicolumn{3}{c} {Macro} \\
    \hline
    Methods 
    & F1 & Rec & Prec 
    & F1 & Rec & Prec\\
    \hline
    \noalign{\smallskip}

    Baseline  
    &	0.4075	&	0.4108	&	0.4043
    &	0.3682	&	0.3778	&	0.3929\\

    OSF
    &	0.4171	&	0.4227	&	0.4116
    &	0.3849	&	0.3798	&	0.4165\\ 

    OSW 
    &	0.4704	&	0.4745	&	0.4664
    &	0.4508	&	0.4375	&	0.4939\\ 

    OSF + OSW 
    &\bf0.4487	&\bf0.4555	&\bf0.4420
    &\bf0.3704	&\bf0.3672	&\bf0.4081\\ 
    \hline
    \end{tabular}
    \end{subtable}
    
    \vskip 10pt
    \begin{subtable}{\textwidth}\centering
    \caption{Recognition - CASME II} 
    \begin{tabular}{lccccccccccccccc}
    \noalign{\smallskip}
    \hline
    Methods 
    & F1 & Rec & Prec\\
    \hline
    \noalign{\smallskip}
	Baseline  
    &	0.5945	&	0.5781	&	0.6118\\ 

    OSF
    &	0.4397	&	0.4109	&	0.4729\\ 

    OSW 
    &	0.6249	&	0.6105	&	0.6399\\ 

    OSF + OSW 
    &\bf0.5869	&\bf0.5696	&\bf0.6053\\ 
    
    \hline
    \end{tabular}
    \end{subtable}
  \end{center}
  \end{table*}
\setlength{\tabcolsep}{1.4pt}
\setlength{\tabcolsep}{4pt}

\setlength{\tabcolsep}{4pt}
\begin{table*}[!tb]
\begin{center}
\caption{F-measure, recall and precision scores for detection and recognition performance on SMIC and CASME II database with LBP-TOP of $8\times 8$ block partitioning}
\label{table:8x8_f_r_p}
   \begin{subtable}{\textwidth}\centering
   \caption{Detection - SMIC} 
    \begin{tabular}{lccccccccccccccc}

    \noalign{\smallskip}
    \hline
    & \multicolumn{3}{c} {Micro} & \multicolumn{3}{c} {Macro}  \\
    \hline
    Methods 
    & F1 & Rec & Prec
    & F1 & Rec & Prec\\
    \hline
    \noalign{\smallskip}

    Baseline  
    &	0.5732	&	0.5732	&	0.5733
    &	0.6005	&	0.5935	&	0.6219\\

    OSF
    &	0.6621	&	0.6616	&	0.6626
    &	0.6729	&	0.6634	&	0.6906\\ 
    
    OSW 
    &	0.6281	&	0.6280	&	0.6281
    &	0.6499	&	0.6337	&	0.6688\\ 

    OSF + OSW 
    &\bf0.7296	&\bf0.7287	&\bf0.7306
    &\bf0.7687	&\bf0.7416	&\bf0.8006\\ 
	
    \hline
 
    \end{tabular}
    \end{subtable}
    
    \vskip 10pt
    \begin{subtable}{\textwidth}\centering
    \caption{Recognition - SMIC} 
    \begin{tabular}{lccccccccccccccc}
    \hline
    & \multicolumn{3}{c} {Micro} & \multicolumn{3}{c} {Macro}  \\
    \hline
    Methods 
    & F1 & Rec & Prec 
    & F1 & Rec & Prec\\
    \hline
    \noalign{\smallskip}

    Baseline  
    &	0.4600	&	0.4613	&	0.4587
    &	0.4111	&	0.4265	&	0.4161\\ 

    OSF
    &	0.4171	&	0.4227	&	0.4116
    &	0.3849	&	0.3798	&	0.4165\\ 

    OSW 
    &	0.5023	&	0.5053	&	0.4993
    &	0.4187	&	0.4276	&	0.4283\\ 

    OSF + OSW 
    &\bf0.5341	&\bf0.5406	&\bf0.5278
    &\bf0.4616	&\bf0.4626	&\bf0.4810\\ 
	
    \hline
    \end{tabular}
    \end{subtable}
    
    \vskip 10pt
    \begin{subtable}{\textwidth}\centering
    \caption{Recognition - CASME II} 
    \begin{tabular}{lccccccccccccccc}
    \hline
    Methods 
    & F1 & Rec & Prec\\
    \hline
    \noalign{\smallskip}

    Baseline  
    &	0.5775	&	0.5610	&	0.5950\\

    OSF
    &	0.4397	&	0.4109	&	0.4729\\

    OSW 
    &	0.5928	&	0.5737	&	0.6132\\ 

    OSF + OSW 
    &\bf0.5719	&\bf0.5582	&\bf0.5862\\ 
	
    \hline
    \end{tabular}
    \end{subtable}
    \end{center}
    \end{table*}
\setlength{\tabcolsep}{1.4pt}
\setlength{\tabcolsep}{4pt}

\subsection{Discussions}
\label{sec:discussion}
In a nutshell, optical strain characterizes the relative amount of displacement by a moving object within a time interval. Its ability to compute any small muscular movements on faces can be advantageous to subtle expression research. By simple product of the LBP-TOP histogram bins with the weights, the resulting feature histograms are intuitively scaled to accommodate the importance of block regions. The $\bf OSF+OSW$ approach generated consistently promising results throughout the experiments tested on the SMIC database. The reason why the proposed method did not performed as excellent on the CASME II as the experiments conducted on SMIC, is probably due to the high frame rate of its acquired video clips. Repetitive frames with very little changes in movements might result in redundancy of input data. Hence, the extracted strain information may be too insignificant (hence negligible) to offer much discrimination between features of different classes. This is most obvious in the $\bf OSF$ results for CASME II, where there was in fact a significant deterioration of performance. SMIC videos, on the other hand (at only halved the frame rate of CASME II videos), is able to 
harness the full capability of optical strain information where the $\bf OSF$ is seen to complement $\bf OSW$ very well, producing even better results when combined together.

Since there are existing background noises in the video frames from both databases, spatial pooling helps to improve the robustness against these noises. Furthermore, high strain magnitudes detected in the frame that exceeded the upper threshold (Eq. \ref{eq:upper-threshold}) are treated as noises and not micro-expression movements. On the other hand, low strain magnitudes below the lower threshold (Eq. \ref{eq:threshold}) will be ignored since they do not contribute sufficient details towards the micro-expressions.

\begin{figure}[t!]
\centering
\includegraphics[width=95mm]{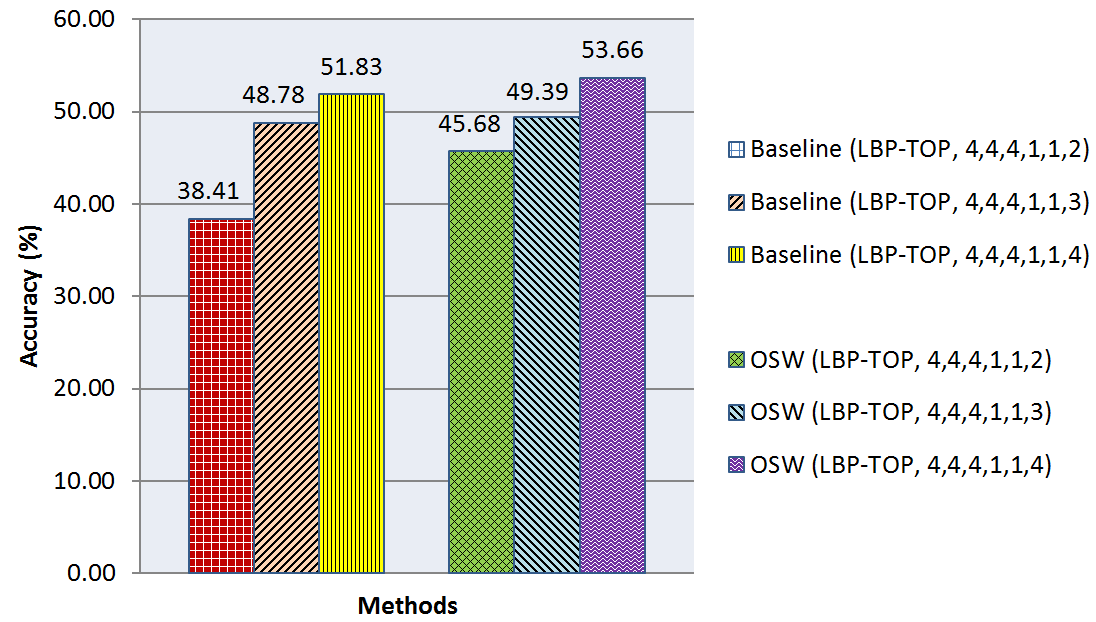}
\caption{Micro-averaged accuracy results of the baseline (LBP-TOP) and $\bf OSW$ methods using different LBP-TOP radii parameters on SMIC database based on LOSOCV 
}
\label{fig:112_113_114}
\end{figure}

Another notable observation worth mentioning lies with the radii parameters of the LBP-TOP feature extractor (which is used by $\bf OSW$ method). By varying the value of $R_T$ (temporal radius), we can observe its importance in the results shown in Fig.~\ref{fig:112_113_114}. The recognition accuracy is the highest for both the $\bf OSW$ and baseline (LBP-TOP) methods when $R_T=4$. Therefore, all the $\bf OSW$ \textcolor{black}{experiments on the SMIC database were conducted using} $\text{\it LBP-TOP}_{4,4,4,1,1,4}$ to maximize the performance on accuracy.
%
%

\begin{figure}[t!]
\centering
\includegraphics[width=0.65\linewidth]{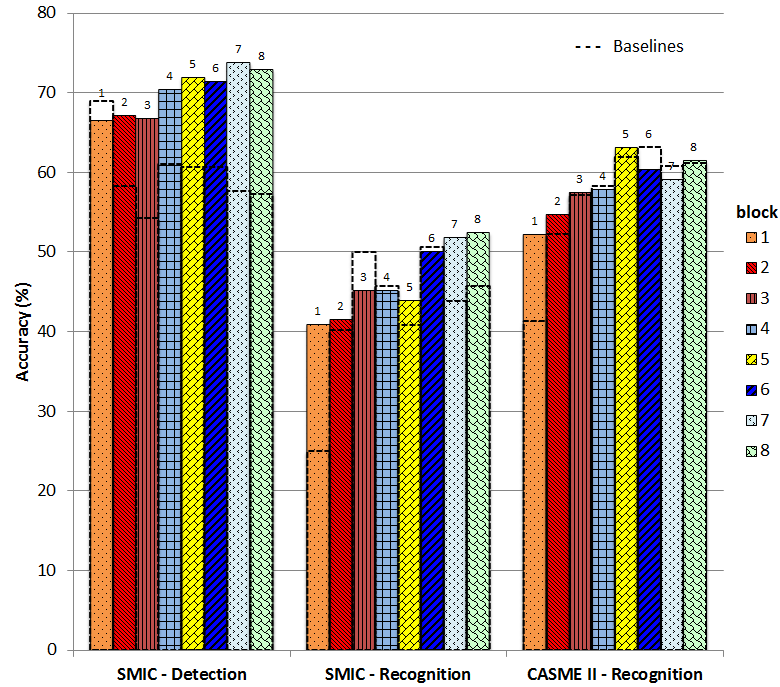}
\caption{Recognition accuracy results of the baseline (LBP-TOP) and $\bf OSF+OSW $ methods using different block partitions in LBP-TOP. The baseline results are denoted by the dashed lines.
}
\label{fig:diff_block}
\end{figure}

We apply settings that were used in the original papers of CASME II~\citep{casme2} and SMIC~\citep{smic}. The LBP-TOP block partitions employed for the detection task in SMIC is $5 \times 5$ blocks, while the recognition tasks in SMIC and CASME II are $8\times 8$ and $5\times 5$ blocks respectively. Fig.~\ref{fig:diff_block} demonstrates in detail how the $\bf OSF+OSW$ fare 
with different block size configuration. with the baselines indicated by dashed lines. 
It can be seen that the larger blocks (i.e. smaller number of partitions, $N=1,2,3$) did not produce better results compared to smaller blocks (i.e. larger number of partitions, $N=6,7,8$) in all scenarios. This is because the local facial appearance and motion that carry important details at specific facial locations are not well described in large block areas. 
Hence, this analysis justifies our choice of using the block settings suggested in the original works, where the best results using the block-based LBP-TOP feature can be achieved. On the other hand, it is also clear in Fig.~\ref{fig:diff_block} that the proposed method, $\bf OSF+OSW$ outperformed the baseline LBP-TOP (dashed lines) in a majority of the experiments.

\subsection{Comparison with other spatio-temporal features}

We compare the results obtained using $\bf OSF+OSW$ against other spatio-temporal based features, namely the optical flow based features $\bf OFF+OFW$, which we construct in the similar manner (optical  flow magnitudes used instead of optical strain magnitudes), \textbf{STIP (HOG+HOF)} or Histogram of Oriented Gradients and Histogram of Optical Flow extracted from spatio-temporal interest points) \citep{wang2009evaluation} and \textbf{HOG3D} or 3D Oriented Gradients \citep{klaser2008hog}. The last two descriptors are popular spatio-temporal features used in various human action recognition~\citep{kovashka2010learning} and facial expression recognition analysis~\citep{hayat2012evaluation}. For both these methods, interest points were densely sampled with their default parameters specified by the authors, and bag-of-words (BOW) \citep{wang2009evaluation} representation is used to learn the visual vocabulary and build the feature vectors. The number of clusters or ``bags" used in the vocabulary learning is determined empirically and the best result is reported. For all these methods, we apply SVM classifier with linear kernel for fair comparison, \christy{except for the method TICS \citep{wang2015micro} and MDMO \citep{liu2016main}, where they classified the micro-expression in CASME II into four categories (i.e., negative, positive, surprise and others), instead of five (i.e., disgust, happiness, tense, surprise and repression). Besides, MDMO utilized polynomial kernel in SVM with heuristic determined parameter settings. }

The recognition accuracy for detection and recognition tasks are reported in Table~\ref{table:comparison}. We observe that STIP and HOG3D features yielded poor results because they are not designed to capture fine appearance and motion changes. The performance of OFF+OFW features are more comparable to the baseline performance of the SMIC, but it is poorer than the CASME II baseline by a significant amount. Overall, the proposed $\bf OSF+OSW$ features yielded promising detection and recognition results compared to the other spatio-temporal features evaluated. We are able to conclude that the proposed method is capable of describing the spatio-temporal information in micro-expressions in a more effective manner.

\setlength{\tabcolsep}{4pt}
\begin{table*}[tb]
\begin{center}
\caption{Comparison of micro-expression detection and recognition performance on the SMIC and CASME II databases for different feature extraction methods}
\label{table:comparison}
\begin{tabular}{lccccc}
\noalign{\smallskip}
\hline
\multirow{2}{*}{Methods} & \multicolumn{2}{c}{$\text{Det-SMIC}^\ast$} & \multicolumn{2}{c}{$\text{Recog-SMIC}^\ast$} & \multirow{2}{*}{\parbox{2cm}{$\text{Recog-}\\ \text{CASMEII}^\bullet$}} \\ 
\cline{2-5}
& Micro & Macro & Micro & Macro \\
\noalign{\vskip 1mm}
\hline
\noalign{\vskip 0.5mm}
 $\text{Baselines}^{\dagger}$
&	60.67	&	66.23
&	45.73	&	47.76
&	61.94\\
\hdashline
\noalign{\vskip 1.5mm} 
STIP \citep{wang2009evaluation}
&	54.88	&	58.53
&	42.07	&	41.42
&	46.96\\ 
HOG3D \citep{klaser2008hog}
&	60.67	&	63.25
&	41.46	&	36.92
&	51.42\\  
\hdashline
\noalign{\vskip 1.5mm} 
RW \citep{oh2015monogenic}
&	N/A	&	N/A 
&	34.15	&	N/A 
&	N/A\\

STM \citep{le2014spontaneous}
&	N/A	&	N/A 
&	44.34	&	N/A 
&	N/A\\

LBP-SIP \citep{wang2015lbp}
&	N/A	&	N/A 
&	54.88	&	N/A  
&	67.21\\


STLBP-IP \citep{huang2015facial}
&	N/A	&	N/A 
&	59.51	&	N/A  
&	N/A \\

$\text{TICS}^\blacklozenge$ \citep{wang2015micro}
&	N/A	&	N/A 
&	N/A	&	N/A 
&	62.30\\

$\text{MDMO}^\blacklozenge$ \citep{liu2016main}
&	N/A	&	N/A 
&	N/A	&	N/A 
&	70.34\\
\hdashline
\noalign{\vskip 1.5mm}  
OSF \citep{liong2014optical}
&	66.16	&	66.34
&	41.46	&	46.00
&	51.01\\ 

OSW \citep{cv4ac}
&	62.80	&	63.37
&	49.39	&	50.71
&	61.94\\ 

OFF + OFW 
&	61.59	&	61.40
&	40.24	&	41.94
&	55.87\\ 

OSF + OSW 
&\bf72.87	&\bf74.16
&\bf52.44	&\bf58.15
&\bf63.16\\

\hline
\footnotesize $\dagger$\hspace{1em}Baseline results from \citep{smic,casme2}\\
\footnotesize $\blacklozenge$\hspace{1em}Used 4 classes for CASME II instead of 5\\
\footnotesize $\ast$\hspace{1em}LOSO cross-validation\\
\footnotesize $\bullet$\hspace{1em}LOVO cross-validation
\end{tabular}
\end{center}
\end{table*}
\setlength{\tabcolsep}{4pt}


\section{Conclusion}
\label{sec:conclusion}
A novel feature extraction approach is proposed for the detection and recognition of facial micro-expressions in video clips. The proposed method describes the fine subtle movements on the face using optical strain technique in two different ways. The first, a direct usage of optical strain information as a feature histogram, and second, the usage of strain information as weighted coefficients to LBP-TOP features. The concatenation of the two feature histograms enable us to achieve promising results in both detection and recognition tasks. Experiments were performed on two recent state-of-the-art databases -- SMIC and CASME II. 

The best detection performance for  SMIC was $74.52\%$ using $5\times 5$ block partition in LBP-TOP, a significant increase of more than $8\%$ compared to the baseline results. The improvement is more significant when $8\times 8$ block partitions were used, where the proposed method was able to achieve a maximum improvement of  $\sim15\%$. In addition, the proposed method was able to achieve an improvement of $+5\%$ and $+10\%$ on micro-expression recognition performance for the SMIC dataset using $5 \times 5$ and $8 \times 8$ block partitions, respectively. The aforementioned results tested on the SMIC dataset were the macro-averaged performance by using SVM with linear kernel. On the other hand, results of the proposed method on the CASME II were slightly better when compared to that of the baselines, i.e., $+0.41\%$ and $+1.22\%$ in $5 \times 5$ and $8 \times 8$ block partitions respectively.

There are many avenues for further research. The kernel function used in SVM is quite sensitive towards the given data, and how this can be better chosen can be further studied. 
Also, better noise filtering techniques and masking of different face regions can be applied to alleviate the instability of illumination and intensity changes on the face areas or background. This can help to reduce the erroneous  optical flow/strain computation.

\section*{Acknowledgment}
This work was funded by Telekom Malaysia (TM) under project 2beAware and by University Malaya Research Collaboration Grant (Title: Realistic Video-Based Assistive Living, Grant Number: CG009-2013) under the purview of University of Malaya Research.

\section*{References}
\bibliography{mybibfile}

\begin{thebibliography}{10}
\expandafter\ifx\csname url\endcsname\relax
  \def\url#1{\texttt{#1}}\fi
\expandafter\ifx\csname urlprefix\endcsname\relax\def\urlprefix{URL }\fi
\expandafter\ifx\csname href\endcsname\relax
  \def\href#1#2{#2} \def\path#1{#1}\fi

\bibitem{nonverbal}
P.~Ekman, W.~V. Friesen, Nonverbal leakage and clues to deception, Journal for
  the Study of Interpersonal Processes 32 (1969) 88--106.

\bibitem{duration}
S.~Porter, L.~ten Brinke, Reading between the lies identifying concealed and
  falsified emotions in universal facial expressions, Psychological Science
  19.5 (2008) 508--514.

\bibitem{six}
P.~Ekman, W.~V. Friesen, Constants across cultures in the face and emotion,
  Journal of personality and social psychology 17(2) (1971) 124.

\bibitem{feel}
M.~G. Frank, M.~Herbasz, K.~Sinuk, A.~Keller, A.~Kurylo, C.~Nolan, I see how
  you feel: Training laypeople and professionals to recognize fleeting
  emotions, in: Annual meeting of the International Communication Association,
  Sheraton New York, New York City, NY, 2009.

\bibitem{security}
M.~G. Frank, C.~J. Maccario, V.~Govindaraju, Protecting Airline Passengers in
  the Age of Terrorism, ABC-CLIO, 2009, pp. 86–--106.

\bibitem{police}
M.~O’Sullivan, M.~G. Frank, C.~M. Hurley, J.~Tiwana, Police lie detection
  accuracy: The eﬀect of lie scenario, Law and Human Behavior 33.6 (2009)
  530–--538.

\bibitem{ucar2016new}
A.~Uçar, Y.~Demir, C.~Güzeliş, A new facial expression recognition based on
  curvelet transform and online sequential extreme learning machine initialized
  with spherical clustering, Neural Computing and Applications 27(1) (2016)
  131--142.

\bibitem{jia2015multi}
Q.~Jia, X.~Gao, H.~Guo, Z.~Luo, Y.~Wang, Multi-layer sparse representation for
  weighted lbp-patches based facial expression recognition, Sensors 15(3)
  (2015) 6719--6739.

\bibitem{zeng2006spon}
Z.~Zeng, Y.~Fu, G.~I. Roisman, Z.~Wen, Y.~Hu, T.~S. Huang, Spontaneous
  emotional facial expression detection, Journal of multimedia 1(5) (2006)
  1--8.

\bibitem{ghi2016facial}
D.~Ghimire, S.~Jeong, J.~Lee, S.~H. Park, Facial expression recognition based
  on local region specific features and support vector machines, Multimedia
  Tools and Applications 16(1) (2016) 1--19.

\bibitem{wang2016facial}
Z.~Wang, Q.~Ruan, G.~An, Facial expression recognition using sparse local
  fisher discriminant analysis, Neurocomputing 174 (2016) 756--766.

\bibitem{liu2016main}
Y.~J. Liu, J.~K. Zhang, W.~J. Yan, S.~J. Wang, G.~Zhao, X.~Fu, A main
  directional mean optical flow feature for spontaneous micro-expression
  recognition, IEEE Transactions on Affective Computing To appear.

\bibitem{wang2015micro}
S.~Wang, W.~Yan, X.~Li, G.~Zhao, C.~Zhou, X.~Fu, M.~Yang, J.~Tao,
  Micro-expression recognition using color spaces, IEEE Transactions on Image
  Processing 24(12) (2015) 6034--6047.

\bibitem{wang2014micro}
S.~J. Wang, W.~J. Yan, X.~Li, G.~Zhao, X.~Fu, Micro-expression recognition
  using dynamic textures on tensor independent color space, in: International
  Conference on Pattern Recognition (ICPR), 2014, pp. 4678--4683.

\bibitem{wang2014face}
S.~J. Wang, H.~L. Chen, W.~J. Yan, Y.~H. Chen, X.~Fu, Face recognition and
  micro-expression recognition based on discriminant tensor subspace analysis
  plus extreme learning machine, Neural Processing Letters 39(1) (2014) 25--43.

\bibitem{huang2015facial}
X.~Huang, S.~J. Wang, G.~Zhao, M.~Piteikainen, Facial micro-expression
  recognition using spatiotemporal local binary pattern with integral
  projection, in: Computer Vision Workshops, 2015, pp. 1--9.

\bibitem{cv4ac}
S.~T. Liong, J.~See, R.~C.-W. Phan, A.~C. Le~Ngo, Y.~H. Oh, K.~Wong, Subtle
  expression recognition using optical strain weighted features, in: ACCV
  Workshops on Computer Vision for Affective Computing (CV4AC), 2014, pp.
  644--657.

\bibitem{casme2}
W.-J. Yan, S.-J. Wang, G.~Zhao, X.~Li, Y.-J. Liu, Y.-H. Chen, X.~Fu, {CASME}
  {II}: An improved spontaneous micro-expression database and the baseline
  evaluation, PLoS ONE 9 (2014) e86041.

\bibitem{smic}
X.~Li, T.~Pfister, X.~Huang, G.~Zhao, M.~Pietikainen, A spontaneous
  micro-expression database: Inducement, collection and baseline, in: Automatic
  Face and Gesture Recognition, 2013, pp. 1--6.

\bibitem{liong2014optical}
S.~T. Liong, R.~C.~W. Phan, J.~See, Y.~H. Oh, K.~Wong, Optical strain based
  recognition of subtle emotions, in: Intelligent Signal Processing and
  Communication Systems (ISPACS), 2014, pp. 180--184.

\bibitem{robust}
M.~J. Black, P.~Anandan, The robust estimation of multiple motions: Parametric
  and piecewise-smooth flow fields, Computer Vision and Image Understanding
  63.1 (1996) 75--104.

\bibitem{towards}
M.~Shreve, S.~Godavarthy, V.~Manohar, D.~Goldgof, S.~Sarkar, Towards macro-and
  micro-expression spotting in video using strain patterns, in: Applications of
  Computer Vision (WACV), 2009, pp. 1--6.

\bibitem{longvideo}
M.~Shreve, S.~Godavarthy, D.~Goldgof, S.~Sarkar, Macro-and micro-expression
  spotting in long videos using spatio- temporal strain, in: Automatic Face,
  Gesture Recognition and Workshops, 2011, pp. 51--56.

\bibitem{dynamic}
G.~Zhao, M.~Pietikainen, Dynamic texture recognition using local binary
  patterns with an application to facial expressions, Pattern Analysis and
  Machine Intelligence, IEEE Transactions 29(6) (2007) 915--928.

\bibitem{davison2014micro}
A.~K. Davison, M.~H. Yap, N.~Costen, K.~Tan, C.~Lansley, D.~Leightley,
  Micro-facial movements: An investigation on spatio-temporal descriptors, in:
  Computer Vision-ECCV Workshops, 2014, pp. 111--123.

\bibitem{wang2015efficient}
Y.~Wang, J.~See, R.~C.~W. Phan, Y.~H. Oh, Efficient spatio-temporal local
  binary patterns for spontaneous facial micro-expression recognition, PloS ONE
  10~(5) (2015) e0124674.

\bibitem{sift}
D.~G. Lowe, Distinctive image features from scale-invariant keypoints,
  International Journal of Computer Vision 60(2) (2004) 91--110.

\bibitem{hog}
N.~Dalal, B.~Triggs, Histograms of oriented gradients for human detection, in:
  Computer Vision and Pattern Recognition, Vol.~1, 2005, pp. 886--893.

\bibitem{boureau}
Y.~L. Boureau, J.~Ponce, Y.~LeCun, A theoretical analysis of feature pooling in
  visual recognition, in: Proceedings of the 27th International Conference on
  Machine Learning, 2010, pp. 111--118.

\bibitem{music}
P.~Hamel, S.~Lemieux, Y.~Bengio, D.~Eck, Temporal pooling and multiscale
  learning for automatic annotation and ranking of music audio, in:
  International Society for Music Information Retrieval Conference, 2011, pp.
  729--734.

\bibitem{survey}
C.~Anitha, M.~K. Venkatesha, B.~S. Adiga, A survey on facial expression
  databases, International Journal of Engineering Science and Technology 2(10)
  (2010) 5158--5174.

\bibitem{usf}
W.~J. Yan, S.~J. Wang, Y.~J. Liu, Q.~Wu, X.~Fu, For micro-expression
  recognition: Database and suggestions, Neurocomputing 136 (2014) 82--87.

\bibitem{poli}
S.~Polikovsky, Y.~Kameda, Y.~Ohta, Facial micro-expressions recognition using
  high speed camera and 3d-gradient descriptor, in: Crime Detection and
  Prevention, 2009, pp. 16--16.

\bibitem{life}
P.~Ekman, Emotions revealed, 2003.

\bibitem{york}
T.~Pfister, Li, X., G.~Zhao, M.~Pietikainen, Recognising spontaneous facial
  micro-expressions, in: International Conference on Computer Vision, 2011, pp.
  1449--1456.

\bibitem{dof}
B.~K. Horn, B.~G. Schunck, Determining optical flow, in: International Society
  for Optics and Photonics, 1981, pp. 319--331.

\bibitem{pof}
J.~L. Barron, D.~J. Fleet, S.~S. Beauchemin, Performance of optical flow
  techniques, International Journal of Computer Vision 12.1 (1994) 43--77.

\bibitem{bainbridge1997determine}
A.~Bainbridge-Smith, J.~Lane, R.~G., Determining optical flow using a
  differential method, Image and Vision Computing 15(1) (1997) 11--22.

\bibitem{ogden1997}
R.~W. Ogden, Non-linear elastic deformations, Courier Corporation, 1997.

\bibitem{barcelos2003well}
C.~A.~Z. Barcelos, M.~Boaventura, E.~C. Silva~Jr, A well-balanced flow equation
  for noise removal and edge detection, Image Processing 12(7) (2003) 751--763.

\bibitem{sobel}
W.~Gao, X.~Zhang, L.~Yang, H.~Liu, An improved sobel edge detection, in:
  Computer Science and Information Technology (ICCSIT), Vol.~5, 2010, pp.
  67--71.

\bibitem{sobel2}
M.~Juneja, P.~S. Sandhu, Performance evaluation of edge detection techniques
  for images in spatial domain, International Journal of Computer Theory and
  Engineering 1(5) (2009) 614--621.

\bibitem{happy2015automatic}
S.~L. Happy, A.~Routray, Automatic facial expression recognition using features
  of salient facial patches, Affective Computing 6~(1) (2015) 1--12.

\bibitem{lucey2010ck}
P.~Lucey, J.~F. Cohn, T.~Kanade, J.~Saragih, Z.~Ambadar, I.~Matthews, The
  extended cohn-kanade dataset (ck+): A complete dataset for action unit and
  emotion-specified expression, in: Computer Vision and Pattern Recognition
  Workshops (CVPRW), 2010, pp. 94--101.

\bibitem{lyons1998jaffe}
M.~J. Lyons, S.~Akamatsu, M.~Kamachi, J.~Gyoba, J.~Budynek, The {Japanese}
  female facial expression ({JAFFE}) database.

\bibitem{ekman}
P.~Ekman, Lie catching and microexpressions, The philosophy of deception (2009)
  118--133.

\bibitem{tsoumakas2010mining}
G.~Tsoumakas, I.~Katakis, I.~Vlahavas, Mining multi-label data, in: Data Mining
  and Knowledge Discovery Handbook, Springer, 2010, pp. 667--685.

\bibitem{le2014spontaneous}
A.~C. Le~Ngo, R.~C.~W. Phan, J.~See, Spontaneous subtle expression recognition:
  Imbalanced databases and solutions, in: Asian Conference on Computer Vision,
  2014, pp. 33--48.

\bibitem{wang2009evaluation}
H.~Wang, M.~M. Ullah, A.~Klaser, I.~Laptev, C.~Schmid, Evaluation of local
  spatio-temporal features for action recognition, in: British Machine Vision
  Conference, 2009, pp. 124--1.

\bibitem{klaser2008hog}
A.~Klaser, M.~Marszałek, C.~Schmid, A spatio-temporal descriptor based on
  3d-gradients, in: British Machine Vision Conference, 2008, pp. 275--1.

\bibitem{kovashka2010learning}
A.~Kovashka, K.~Grauman, Learning a hierarchy of discriminative space-time
  neighborhood features for human action recognition, in: Computer Vision and
  Pattern Recognition (CVPR), 2010, pp. 2046--2053.

\bibitem{hayat2012evaluation}
M.~Hayat, M.~Bennamoun, A.~El-Sallam, Evaluation of spatiotemporal detectors
  and descriptors for facial expression recognition, in: Human System
  Interactions (HSI), 2012, pp. 43--47.

\bibitem{oh2015monogenic}
Y.~H. Oh, A.~C. Le~Ngo, J.~See, S.~T. Liong, R.~C.~W. Phan, H.~C. Ling,
  Monogenic riesz wavelet representation for micro-expression recognition, in:
  Digital Signal Processing, 2015, pp. 1237--1241.

\bibitem{wang2015lbp}
Y.~Wang, J.~See, R.~C.~W. Phan, Y.~H. Oh, Lbp with six intersection points:
  Reducing redundant information in lbp-top for micro-expression recognition,
  in: Computer Vision--ACCV, 2015, pp. 525--537.

\end{thebibliography}
\end{document}